\documentclass[twocolumn]{IEEEtran}
\usepackage{booktabs}
\usepackage{url}
\usepackage{caption}
\usepackage{graphicx}
\usepackage{color}
\usepackage{array}
\usepackage{float}
\usepackage[utf8]{inputenc}
\usepackage[export]{adjustbox}
\usepackage{multirow}
\usepackage{amsmath}
\usepackage{tcolorbox}
\usepackage{pdfpages}
\graphicspath{{Images/}}

\begin{document}             

\title{Machine Learning for Predicting Epileptic Seizures Using EEG Signals: A Review}
\author{Khansa Rasheed$^1$, Adnan Qayyum$^1$, Junaid Qadir$^1$, Shobi Sivathamboo$^{2, 3, 4}$, Patrick Kwan$^{2, 3, 4}$, Levin Kuhlmann$^{5, 6}$, Terence O'Brien$^{2, 3, 4}$, and Adeel Razi$^{7, 8, 9, 10}$\\\vspace{2mm}
$^1$ Information Technology University (ITU)-Punjab, Lahore, Pakistan\\
$^2$ Department of Neuroscience, Central Clinical School, Monash University, Melbourne, Victoria, Australia\\
$^3$ Department of Neurology, Alfred Health, Melbourne, Victoria, Australia\\ 
$^4$Departments of Neurology and Medicine, The Royal Melbourne Hospital, The University of Melbourne, Melbourne, Victoria, Australia.\\
$^5$ Faculty of Information Technology, Monash University, Clayton, Australia\\
$^6$ Department of Medicine, St. Vincent’s Hospital, The University of Melbourne, Parkville, Australia\\
$^7$ Turner Institute for Brain and Mental Health, Monash University, Clayton, Australia\\
$^8$ Monash Biomedical Imaging, Monash University, Clayton, Australia\\
$^9$ Wellcome Centre for Human Neuroimaging, UCL, London, United Kingdom\\
$^{10}$ Department of Electronic Engineering, NED University of Engineering and Technology, Karachi, Pakistan\\
}
\maketitle

\begin{abstract}
With the advancement in artificial intelligence (AI) and machine learning (ML) techniques, researchers are striving towards employing these techniques for advancing clinical practice. One of the key objectives in healthcare is the early detection and prediction of disease to timely provide preventive interventions. This is especially the case for epilepsy, which is characterized by recurrent and unpredictable seizures. Patients can be relieved from the adverse consequences of epileptic seizures if it could somehow be predicted in advance. Despite decades of research, seizure prediction remains an unsolved problem. This is likely to remain at least partly because of the inadequate amount of data to resolve the problem. There have been exciting new developments in ML-based algorithms that have the potential to deliver a paradigm shift in the early and accurate prediction of epileptic seizures. Here we provide a comprehensive review of state-of-the-art ML techniques in early prediction of seizures using EEG signals. We will identify the gaps, challenges, and pitfalls in the current research and recommend future directions. 
\end{abstract}

\begin{IEEEkeywords}
Epileptic Seizure, EEG, Machine Learning
\end{IEEEkeywords}

\section{Introduction} 

Epilepsy is a group of neurological disorders that are characterized by an enduring predisposition to generate recurrent seizures and can affect individuals of any age. Epilepsy arises from the gradual neurobiological process of `\textit{epileptogenesis}' \cite{rakhade2009epileptogenesis}, which causes the normal brain network to fire neurons in a self-sustained hyper-synchronized manner in the cerebral cortex. According to the World Health Organization (WHO), 70 million people worldwide have epilepsy and epilepsy trails only migraine, stroke, and Alzheimer's disease in the list of the most widespread brain diseases \cite{england2012epilepsy}. The seizures caused by epilepsy are debilitating and disrupt the day-to-day activities of the patients, and are associated with an increased risk of premature mortality. The dearth of neurologists in many countries complicates the management of epilepsy---especially in the developing countries where the neurologists are in short supply.

Even though epilepsy and seizures are sometimes referred to synonymously in some literature, it is worth noting that not all seizures are epileptic and convulsions and seizures may also occur due to acute neurological insults (such as stroke, brain trauma, metabolic disturbances, and drug toxicity) without necessarily reflecting a long term predisposition to recurrent unprovoked seizures (i.e. epilepsy).

An epileptic seizure (ES) is caused by a sudden abnormal, self-sustained electrical discharge that occurs in the cerebral networks and usually lasts for less than a few minutes. ES attacks are hard to predict, moreover, severity and duration of attack also cannot be anticipated. Therefore, injuries and safety issues from the events are a major concern for patients and their families. Hence, early prediction of epilepsy attacks is crucial to avoid and counter their adverse consequences. The brain activity of patients with epilepsy can be categorized as different states: pre-ictal (immediately preceding seizure), ictal (during a seizure), post-ictal (immediately following a seizure), and interictal (in-between seizures). Further details of these terms are provided in the section of the paper. ES prediction is a classification problem, i.e. differentiating between the pre-ictal and interictal states. Due to the recurrent nature of epilepsy, ES occurs in groups and patients afflicted from seizure clusters can acquire advantage through the forecasting of follow-on seizures.

\begin{table*}[]
\caption{Comparison of this paper with existing surveys. Legends: $\surd$= discussed, $\times$= not discussed, $\approx$ = partially discussed.}
\label{tab:table1}
\centering
\begin{tabular}{ l l l l l l l l l }
\hline
\textbf{\begin{tabular}[c]{@{}l@{}}Reference\end{tabular}} & \textbf{Year} & \textbf{Focused Area} & \textbf{\begin{tabular}[c]{@{}l@{}}EEG \\ Analysis \\ Techniques\end{tabular}} & \textbf{\begin{tabular}[c]{@{}l@{}}Feat-\\ ures\end{tabular}} & \textbf{ML} & \textbf{DL} & \textbf{Pitfalls} & \textbf{\begin{tabular}[c]{@{}l@{}}Future\\ Direction\end{tabular}} \\ \hline
Mormann et al. \cite{mormann2006seizure} & 2006 & Epilepsy prediction & $\times$ & $\times$ & $\surd$ & $\times$ & $\approx$ & $\surd$\\ \hline
Subha et al. \cite{subha2010eeg} & 2008 & EEG signals & $\surd$ & $\surd$ & $\times$ & $\times$ & $\approx$ & $\times$ \\ \hline
Yuedong et al. \cite{song2011review} & 2011 & EEG signals & $\surd$ & $\approx$ & $\times$ & $\times$ & $\times$ & $\times$ \\ \hline
Acharya et al. \cite{acharya2013automated} & 2013 & \begin{tabular}[c]{@{}c@{}}EEG signals for epilepsy\end{tabular} & $\surd$ & $\surd$ & $\surd$ & $\times$ & $\times$ & $\surd$ \\ \hline
Gadhoumi et al. \cite{gadhoumi2016seizure}& 2016& ES prediction & $\times$ & $\surd$ & $\surd$ & $\times$ & $\surd$ & $\approx$ \\ \hline
M.Iftikhar et al. \cite{iftikhar2018survey} & 2018 & DL for EEG signals & $\surd$ & $\approx$ & $\times$ & $\surd$ & $\times$ & $\times$ \\ \hline
Acharya et al. \cite{acharya2018automated}&2018& ES prediction &$\times$ & $\surd$ & $\surd$ & $\approx$ & $\times$ & $\surd$\\ \hline
Kuhlmann et al. \cite{acharya2018automated}&2018& ES prediction &$\times$ & $\times$ & $\surd$ & $\approx$ & $\times$ & $\surd$\\ \hline
Roy et al. \cite{roy2019deep}& 2019 & DL for EEG & $\surd$ & $\times$ & $\times$ & $\surd$ & $\surd$ & $\surd$ \\ \hline
Li et al. \cite{lideep} & 2019 & DL for EEG & $\times$ & $\times$ & $\approx$ & $\surd$ & $\surd$ & $\surd$\\ \hline
This paper & 2019 & ES prediction & $\surd$ & $\surd$ & $\surd$ & $\surd$ & $\surd$ & $\surd$ \\ \hline
\end{tabular}
\label{tab:com}
\end{table*}
Electroencephalography (EEG) is a particularly effective diagnostic tool to study the functional anatomy of the brain during an ES attack. The prediction and medication of epilepsy have been broadly studied through EEG. EEG signals, which are non-Gaussian and non-stationary, measure the electrical activity in the brain which are in turn used to diagnose the type of the brain disorders. The analysis of EEG measurements helps segregate normal and abnormal function of the brain. For an accurate prediction of epilepsy, it is necessary to examine EEG recordings of longer duration. Expert neurologists examine epilepsy by studying continuous EEG signals recorded over several days, weeks, or even months, which requires a huge amount of human effort and time. Over the years, various studies have employed machine learning (ML)-based prediction methods to address this issue. Deep learning (DL) is an advanced ML technology that is capable of learning patterns more precisely from large collections of data by processing it through a multi-layer hierarchical architecture. The ability of DL to produce very accurate results has influenced the researchers to tackle numerous real-world applications by employing DL techniques with various researchers proposing DL-based approaches for the ES prediction in the last five to six years. 



The objective of this paper is to accentuate the primary advances in the employment of ML methods for epilepsy prediction. We will provide a brief introduction to neuroscience, various tools used for studying brain, and how they have been or could be used for the prediction of epilepsy.

\vspace{2mm}
\textit{Contributions of this paper}:
Although there exist several reviews that specifically cover epilepsy seizure prediction using EEG signals, to the best of our knowledge, there does not yet exist a review that covers in depth the application of ML methods for predicting epileptic seizures. For instance, Mormann et al. have provided an overview of the evolution of seizure predicting methods since the 1970s till 2006 \cite{mormann2006seizure} and have covered the major issues related to methodology of ES prediction. Gadhoumi et al. have provided a brief overview of valid methods used for ES prediction and comprehensively described the statistical significance of the results of the prediction \cite{gadhoumi2016seizure}. In the recently published review, Kuhlmann et al. have briefly described the advancement in the field of ES prediction and ES prediction competitions. They concluded that these advancements with standard statistical evaluations are opening ways for the development of ES prediction methodologies and they refined the existing guidelines to achieve this development \cite{kuhlmann2018seizure}. This survey is unique because it provides comprehensive answers to questions like why there is a need for ML techniques for ES prediction, how the evolution of relatively newer techniques like DL is proving highly useful for ES prediction, and discusses directions for future research in this area. The comparison of this paper with existing surveys is presented in Table \ref{tab:com}. 
\begin{table}[]
\centering
\caption{List of Acronyms}
\label{tab:3}
\begin{tabular}{|l|l|}
\hline
ANN & Artificial Neural Network \\ \hline
ApEn & Approximate Entropy \\ \hline
BLDA & Bayesian Linear Discriminant Analysis \\ \hline
CNN & Convolutional Neural Network \\ \hline
CWT & Continuous Wavelet Transform \\ \hline
DNN & Deep Neural Network \\ \hline
DWT & Discrete Wavelet Transform \\ \hline
EMD & Empirical Mode Decomposition \\ \hline
ES & Epileptic Seizure \\ \hline
EEG & Electroencephalography \\ \hline
FD & Fractal Dimension \\ \hline
FPR & False Prediction Rate \\ \hline
FT & Fourier Transform \\ \hline
HE & Hurst Exponent \\ \hline
HHT & Hilbert-Huang Transform \\ \hline
HP & Hjorth Parameter \\ \hline
LLE & Largest Lyapunov Exponent \\ \hline
LSTM & Long-Short Term Memory \\ \hline
MLP & Multi-layer Perceptron \\ \hline
RNN & Recurrent Neural Network \\ \hline
SEF & Spectral Edge Frequency \\ \hline
SBP & Spectral Band Power \\ \hline
SM & Statistical Moment\\ \hline
SOM & Self Organizing Map \\ \hline
SpM & Spectral Moment \\ \hline
TPR & True Positive Rate \\ \hline
WFT & Wavelet Fourier Transform \\ \hline
WT & Wavelet Transform \\ \hline

\end{tabular}
\end{table}

\begin{figure*}[!htb]
\centering
\includegraphics[width=0.9\textwidth]{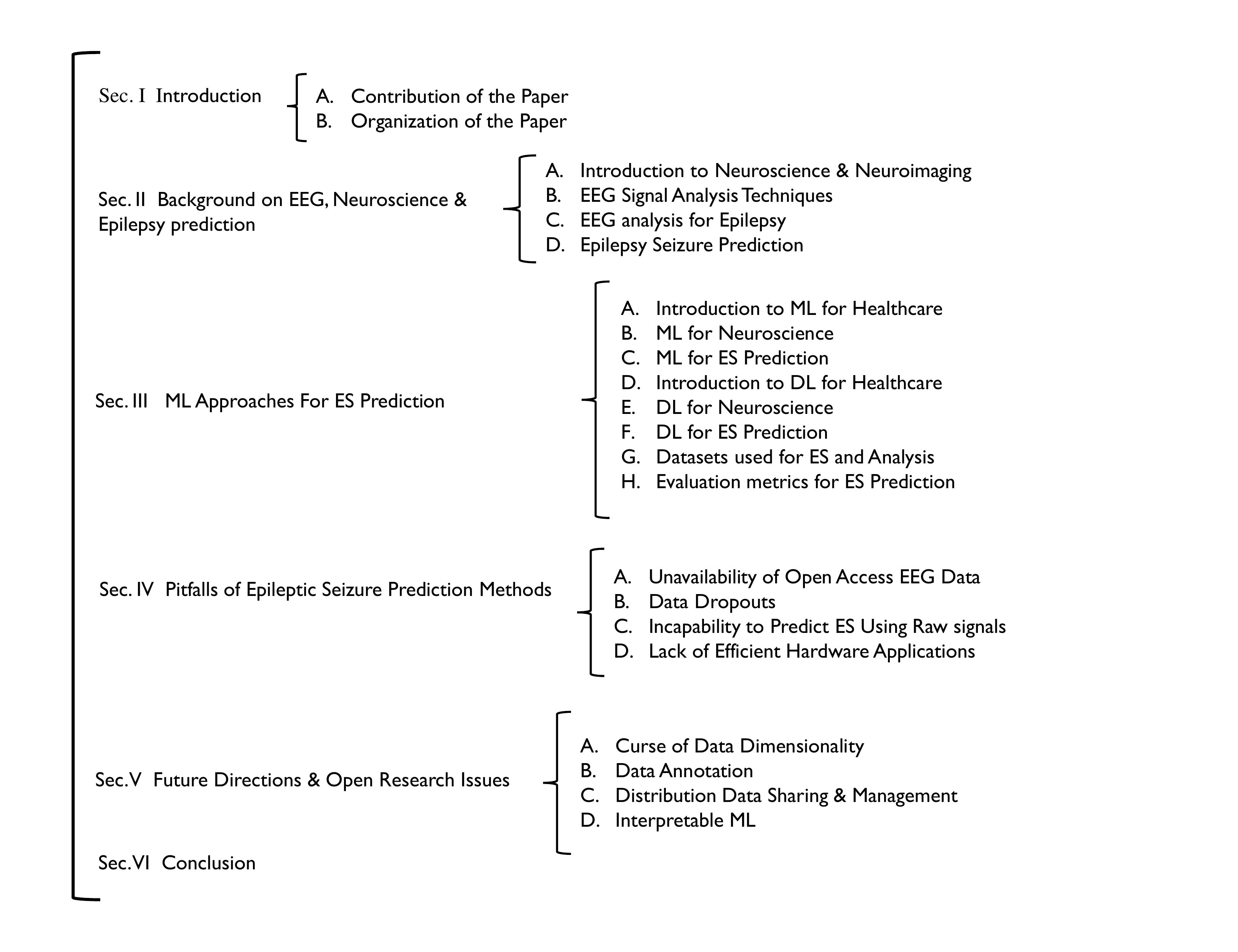}
\caption{Organization of the paper}
\label{fig:org}
\end{figure*}

\textit{Organization of this paper}:
The organization of this paper is depicted in Figure \ref{fig:org}. In Section \ref{sec:back}, a brief background of neuroscience, EEG, and epilepsy prediction is presented. Section \ref{sec:ML_ES} covers data-driven ML approaches for ES prediction. Section \ref{sec:pitfalls} comprises of identifying several pitfalls in applying these methods. Future directions and open research problems are presented in Section \ref{sec:insights}. Finally, the paper is concluded in Section \ref{sec:con}. List of acronyms used in the paper is provided in Table \ref{tab:3}.

\section{Background on EEG, Neuroscience, and Epilepsy Prediction}
\label{sec:back}

\subsection{A Brief Introduction to Neuroscience and Neuroimaging}

Neuroscience is the multidisciplinary study of the brain. It integrates multifarious disciplines including \textit{neuroanatomy} (in which neuroanatomists engage with the structures of the human brain), \textit{neurochemistry} (where chemists observe the chemical properties of intercommunication in the brain), \textit{neurophysiology} (where the neurophysiologists investigate the electrical properties of the brain) and \textit{neuropsychology} (where psychologists endeavor to interpret the cognitive domains and the structures that sustain those cognitive domains in neuroscience) \cite{weyhenmeyer2006rapid}. Neuroscience also has further divisions for e.g, molecular neuroscience, cognitive neuroscience \cite{lieberman2000intuition}, clinical neuroscience, computational neuroscience \cite{sejnowski1988computational}, developmental neuroscience, and cultural neuroscience, to name just a few.

\begin{table*}[]
\centering
\caption{EEG frequency bands and related study of brain functioning.}
\begin{tabular}{|c|c|l|}
\hline
\textbf{Frequency Bands} & \textbf{Frequency Range (Hz)} & \multicolumn{1}{c|}{\textbf{Relation to human behavior}}          \\ \hline
\textit{Delta}   & 1-4       & Predominantly found in infants and deep sleep stages of normal adults   \\ \hline
\textit{Theta}   & 4-8       & High value of theta rhyme in awake adults show the abnormal cognitive activity \\ \hline
\textit{Alpha}   & 8-12       & Usually present in posterior region of the brain and in normal relaxed adults \\ \hline
\textit{Beta}   & 12-26       & Present in frontal region of the brain and in alert anxious person    \\ \hline
\textit{Gamma}   & 26-1000      & Predominantly found in stressed, happy or aware person       \\ \hline
\end{tabular}

\label{tab:freq}
\end{table*}

The brain is anatomically segregated to communities which make up a functionally specialized brain network (functional segregation). These functionally segregated communities are functionally interconnected (functional integration) to perform very complex tasks like they implement cognition \cite{razi2016connected}. Neuroimaging uses various ways to directly or indirectly image the structure and the function of the central nervous system. Two broad categories are structural imaging that pertains to anatomy, pathology or injury and functional imaging that deals with metabolism, pharmacology or cognition. Some of the important and widely used neuroimaging techniques are namely: computed tomography (CT) that computes the absorbed amount of X-rays to provide a series of cross-sectional images of the brain; positron emission tomography (PET) that generates the image of active molecule binding; structural magnetic resonance imaging (MRI) that examines the anatomy and pathology of the brain; functional MRI that examines the brain activity; diffusion MRI that maps the diffusion of water molecules in the brain to reveal macroscopic details of brain tissues; and magnetic resonance spectroscopy (MRS) imaging is used to study the metabolic changes in brain tumors, stroke, and seizure, etc. The brain's electrical activity in different physiological situations can be measured using EEG, which falls under the category of functional imaging.

\subsection{EEG Signal Analysis Techniques}

\subsubsection{Introduction to EEG}

In 1923, Hans Berger contrived EEG, a non-invasive functional imaging methodology to study the brain. EEG records the electrical signals from the cerebral cortex by measuring the electrical activity of the group of neurons. Compared to the functional MRI, EEG provides a higher temporal insight into neural activity but has a lower spatial resolution. Typically, five frequency bands are analysed for processing EEG signals, \textit{Delta} (up to 4 Hz), \textit{Theta} (4--8 Hz), \textit{Alpha} (8--12 Hz), \textit{Beta} (12--26 Hz), and \textit{Gamma} (26--100 Hz). A summary of these frequency bands with their relation to human behavior is presented in Table \ref{tab:freq}, The amplitude of EEG range from 10 $\mu$V--100 $\mu$V while its frequency ranges from 1 Hz--100 Hz. 

To diagnose a disease, or to decode brain activity by using EEG data, one initially extracts features or uses spectral information of raw EEG data by applying Fourier transform (FT) or wavelet transform (WT). These extracted features or transformed raw data is then used to train an ML-based classifier while DL algorithms have been proved efficient for automatic extraction of feature for training.

There are two methods of EEG recording based on the position of the reference electrode.
\begin{enumerate}
\addtolength{\itemindent}{1.0068 cm}
\item[(1)] \textit{Bipolar Montage}: In a bipolar montage, both electrodes are placed on an electrically active region of the scalp and the voltage difference between electrodes is measured.

 \item[(2)] \textit{Monopolar Montage/Unipolar Montage}: In a mono-polar montage, one electrode is placed on an electrically active region and the reference electrode is placed on an electrically inactive region (e.g., an ear lobe).
\end{enumerate}

The traditional method for the recording of EEG signals is to place the electrodes on the surface of the skull, which is known as scalp EEG. The main drawback of scalp EEG is that the recorded signals become distorted owing to a large distance between neurons inside the skull and the electrodes. For the quality of signals to be enhanced in terms of distortion and amplitude, intracranial electroencephalography (iEEG) signals are recorded by placing the electrodes on the exposed surface of the brain.

EEG possesses several characteristics which makes it quite preferable to use for ES prediction research. Along with its ability to track the various changes occurring in the brain during epilepsy, another main feature it provides is the relatively lower hardware cost which makes it able to be used for a large number of patients and to record for longer duration. Multiple other techniques such that fMRI or MEG require bulky and immobile equipment which piles up the cost to millions of dollars. Among the current practical approaches to predict epileptic seizures, \textit{My Seizure Gauge} is the most example of a wearable device created to work as a personalized advisory device for seizure prediction \cite{dumanis2017seizure}. This device can cover intracranial EEG recordings, scalp EEG, electromayography (EMG) (recording of the electrical activity produced by skeletal muscles), electrocardiography (ECG) (recording of the electrical activity of heart), electrodermal activity (EDA), photoplethysmography (PPG), and respiration.

\subsubsection{Analysis Techniques}

EEG analysis methods can mainly be classified into the time domain methods, frequency domain methods, time-frequency domain methods, and linear or non-linear methods \cite{subha2010eeg}.

\paragraph{\textbf{Time domain methods}} 

EEG recordings are non-stationary and non-linear functions of time. \textit{Linear prediction} is a time-domain method in which the output is calculated from the input and earlier outputs. \textit{Principal component analysis (PCA)}, \textit{linear discriminant analysis (LDA)} and \textit{independent component analysis (ICA)} are widely used unsupervised time-domain methods to summarize EEG data. PCA is used to transform the high-dimensional data (in case of epilepsy high-dimensional feature vectors) to a low-dimensional data while ICA decomposes high-dimensional data into linear statistically independent components. In EEG data analysis, ICA is most commonly used to remove artifacts. Whereas, LDA is used to reduce dimensions of feature sets by finding linear combinations of feature vectors.

\paragraph{\textbf{Frequency domain methods}} 

During an epileptic seizure, there is a sudden change in the frequency of EEG signals, which is measurable by applying frequency-domain methods, e.g., using Fourier transform (FT). One can used either \textit{parametric} or \textit{non-parametric methods} to estimate the power spectrum using FT. Welch (a non-parametric) method, a modified version of widely used \textit{periodogram} method, is generally used for the estimation of PSD. But this has a disadvantage of spectral leakage and is overcome by employing parametric methods. Parametric methods provide better frequency resolution by assuming the EEG signal is a stationary random process. Moving average (MA), auto-regression (AR), and auto-regressive moving average (ARMA) are commonly applied parametric methods.

\paragraph{\textbf{Time-frequency domain methods}} Above mentioned time-domain and frequency-domain methods have limitations of providing exact frequencies involved at a particular time instant and the information of time moment respectively. To overcome these limitations, wavelet transform (WT), a time-frequency based analysis technique, is widely used to obtain multi-resolution decomposed sub-band signals by passing the EEG signal through filter banks.

\paragraph{\textbf{Non-linear methods}} Non-linear analysis methods are applied to detect the coupling among harmonics in signal's spectrum. \textit{Higher order spectra (HOS)}, various measures of entropy---e.g., \textit{approximate entropy (ApEn)}; \textit{Kolmogorov entropy}; \textit{sample entropy (SampEn)}---along with the \textit{Hurst exponent (H)}, \textit{largest Lyapunov exponent (LLE)} are widely used non-linear parameters for EEG analysis. Entropy and LLE are commonly used as features for epilepsy classification. Entropy provides clues about information stored in the probability distribution of a signal and measures the uncertainty or randomness in the patterns of the data. A higher value of the entropy refers to highly random patterns of data. LLE provides the information of the dependence of the system on initial conditions. For a more detailed review of the analysis techniques, the interested readers are referred to \cite{acharya2013automated} and \cite{subha2010eeg}.

\subsection{EEG Signal Analysis for Epilepsy}

Analysis of the EEG signals is the primary method to identify ES activities in the brain. EEG recordings are an important clinical tool for distinguishing ES from non-ES. EEG signals recorded, before and during a seizure, contains characteristics that can be used to identify the different stages of an epileptic seizure, and the pre- and post-seizure periods. These stages are briefly described below \cite{acharya2013automated}.

\begin{enumerate}
\item[(a)] \textit{Pre-ictal State:} A pre-ictal state becomes apparent during a said time period before the occurrence of a seizure and does not occur at the rest of the times. It might not necessarily be visually apparent. However, it will reflect changes in the underlying signals and would be predictive of seizures within a specific range of values. For a pre-ictal state to be of use clinically in a warning system, it has to be detected early enough so that the time under false warning is minimized \cite{kuhlmann2018seizure}.

\item[(b)] \textit{Pro-Ictal State:} In this state seizures are more likely but not guaranteed to happen.

\item[(c)] \textit{Ictal and Interictal State:} The ictal state is a change in EEG signals during a seizure and interictal is the stage between two following seizure onsets. For the same person, the number of epileptogenic neurons, cortical region, and the span of seizure can be altered.

\item[(d)] \textit{Post-Ictal State:} This state is after the occurrence of a seizure.
\end{enumerate}

The wave pattern may hold valuable information about brain activity. Experienced neurologists can detect disorders by visually observing the EEG signals. However, this procedure is time-consuming and is prone to faulty detection due to high temporal and spatial aspects of the dynamic non-linear EEG data. Therefore, computerized techniques, EEG signal parameters extraction, and analysis can be profoundly beneficial in the diagnostics. 

\begin{figure}[!htb]
\centering
\includegraphics[width=0.5\textwidth]{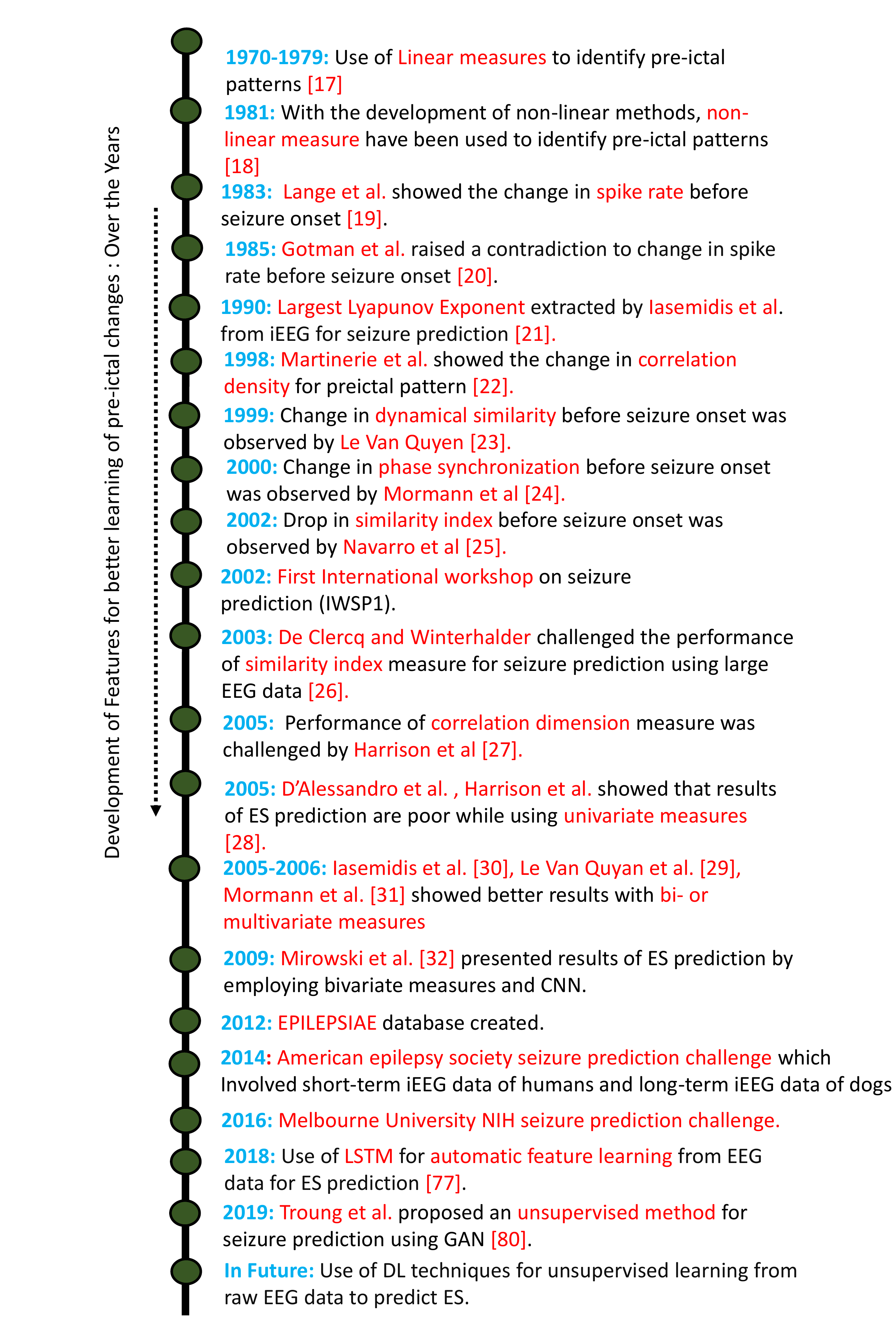}
\caption{ The timeline for the development of EEG measures used for ES prediction.}
\label{fig:timeline}
\end{figure}

\subsection{Epilepsy Seizure Prediction}
In the 1970s, early research of ES prediction carried out using linear approaches of feature extraction \cite{viglione1975proceedings}. While in 1980s, the development of non-linear methods helped researchers to employ these techniques for feature extraction because of the non-linear nature of EEG signals \cite{rogowski1981prediction} \nocite{lange1983temporo} \nocite{gotman1985electroencephalographic} \nocite{iasemidis1990phase} \nocite{martinerie1998epileptic} \nocite{le1999anticipating} \nocite{mormann2000mean} \nocite{navarro2002seizure} \nocite{de2003anticipation} \nocite{harrison2005accumulated} \cite{harrison2005correlation} \nocite{le2005preictal} \nocite{iasemidis2005long} \nocite{chaovalitwongse2005performance} \nocite{mirowski2009classification}. With the recognition of EEG patterns of epilepsy---i.e, pre-ictal, ictal, and interictal patterns---the use of the pre-ictal stage for ES detection was also applied in this decade. In 1998, early prediction of ES almost 6 sec before the seizure onset, was carried out by Salant et al. \cite{salant1998prediction} which was further developed by Drogenlen et al. in 2003 \cite{van2003seizure}. They used Kolmogorov entropy as a feature to predict ES 2--40 min before onset. First international workshop on ES prediction was held in 2002 in which dataset of multi-day recordings of EEG provided by different epilepsy centers. Later, several studies were carried out on this dataset \cite{lehnertz2005first}. In 2003, Mormann et al. used the fact that the hyper-synchronous firing of neurons in the brain is a cause of ES and found that the phase synchronization of different EEG channels decrease before seizure onset \cite{mormann2003epileptic}. In the first decade of the present century, studies based on extensive EEG data have raised doubts about the performance of measures calculated in the previous century. Researchers found that the results of earlier studies based on a selected and inadequate amount of data could not be reproduced on extensive and unseen data.

It was decided to conduct competitions on seizure prediction in international workshops conducted on the said topic. The purpose of these competitions was to standardize the comparison of the performance of algorithms trained on a common dataset. The first seizure prediction competition was held in collaboration with International Workshop on Seizure Prediction 3 (IWSP3) in 2007 while the second competition conducted in 2009 was in collaboration with IWSP4. In both the competitions, the contestants were provided with the continuous iEEG recording from three epileptic patients. However, the performance results of the algorithms were not satisfactory. American Epilepsy Society Seizure Prediction Challenge which was held in 2014, involved short-term human iEEG containing 942 seizures recorded over more than 500 days and long-term iEEG recordings of dogs with epilepsy. All contestants were provided with the same 10 min long training and testing data. The Area Under the Curve (AUC) was used as a performance evaluation metric. With the same structure, another contest held by Melbourne University which involved long-term iEEG recording with 1139 seizures. For more details of the contest see \cite{kuhlmann2018epilepsyecosystem}. The contests were open to any algorithm computing basic features of EEG signals for ES prediction or machine learning models trained on these basic features. In any case we still do not really know what features or algorithms are best. In the contests, people submitted algorithms, that were too complicated. So it is difficult to say which feature or ML algorithm was best. The organizers of the contests are working towards dissecting it now with \url{Epilepsyecosystem.org}. Recent work of Matias Maturana et al. \cite{maturana2019critical} presents a solution that might work well across patients. They identified the critical slowing of brain signals as an indicator for ES prediction.  A timeline for the development of the EEG data measures is depicted in Figure \ref{fig:timeline}, the interested readers can refer to \cite{editor2008seizure} for more detailed information about the history of these developments.

\section{ML Approaches for ES Prediction}
\label{sec:ML_ES}

In this section, we provide a comprehensive review of the literature using ML-based methods for ES prediction and we start this section by first highlighting the potential of using ML techniques for healthcare and neuroscience applications.

\subsection{Introduction to ML for Healthcare}
ML is proliferating across research areas over the past few decades by using statistical methods to recognize patterns in large collections of data. The availability of large-scale biomedical data is turning over a new leaf for healthcare researchers. Development of effective medical tools relies on data analysis approaches and the advancements of ML techniques. Because the manual detection of representations is not possible due to the complex structure of medical data and that is why ML is extensively used in healthcare for the diagnosis of diseases, e.g., detection of breast cancer \cite{kourou2015machine}, classification of skin cancer \cite{jorgensen2008machine}, diagnosis of Alzheimer disease \cite{moradi2015machine}, prediction of epilepsy \cite{yang2018epileptic}, and diagnosis of diabetic retinopathy in retinal images \cite{gulshan2016development}. Electronic health records (EHR)-based ML algorithms have proved beneficial for prediction of future diseases and are capable of automatically diagnosing patients given their clinical status \cite{mazurowski2008training} although still much work is needed. Biomedical fields with large image datasets---such as radiology \cite{zaharchuk2018deep}, cardiology \cite{johnson2018artificial}, pathology \cite{kukharchik2007vocal}, and genomics \cite{libbrecht2015machine}---are using various ML methods for automatic diagnosis, classification, and prediction of various disease.

\subsection{ML for Neuroscience}

Learning about the structure and functional anatomy of the human brain has been the foremost focus of neuroscientists in recent years. The advancements in technology have enabled the neuroscientists to acquire, process, and analyse the neuroimaging data at  unprecedented detail, while ML and DL are the paramount examples for such enabling technologies that can be used as a potential exploratory source for building theories about brain functioning for neuroscientists \cite{hinton2011machine}. In this section, we provide a general introduction to various ML techniques (e.g., \textit{supervised learning}, \textit{unsupervised learning}, and \textit{reinforcement learning}) that have been used in the field of neuroscience. 

\subsubsection{Supervised Learning}

In supervised learning, training data accompanied by labels assigned by human experts is fed to the learning algorithm for extracting the relation between data and labels so that the system can classify the unseen data accurately to their respective categories. For instance, a training data consists of images with labels of house, a dog, a cat and we want an algorithm that can predict the label of an image previously unknown to the system. These algorithms have wide applications in the field of computational and theoretical neuroscience---an example technique is support vector machine (SVM), a supervised learning algorithm generally used for prediction of ES (described in a latter section). Analysis of neural mechanisms under stress is carried out using a supervised ML approach \cite{vu2018shared}. 

\subsubsection{Unsupervised Learning}
Our brain receives most of the information in a day without any guidance. The brain develops a working model from the repetition of information and uses this model to make a perception. This perception is then used for detecting the patterns in new information. Unsupervised learning algorithms are motivated by how the brain studies new things through perceptions. Unsupervised learning applies unclassified or unlabeled data for training of the algorithms. These algorithms are extensively used in the identification and classification of diseases from neurophysiological data. As a representative example, we refer to the work of Drysdale et al. who classified depression types using fMRI \cite{drysdale2017resting} and the work of O'Donnell et al. who used a clustering algorithm for the identification of white matter tracts from diffusion MRI \cite{o2006method}.

\begin{figure*}[!htb]
\centering
\includegraphics[width=0.9\textwidth]{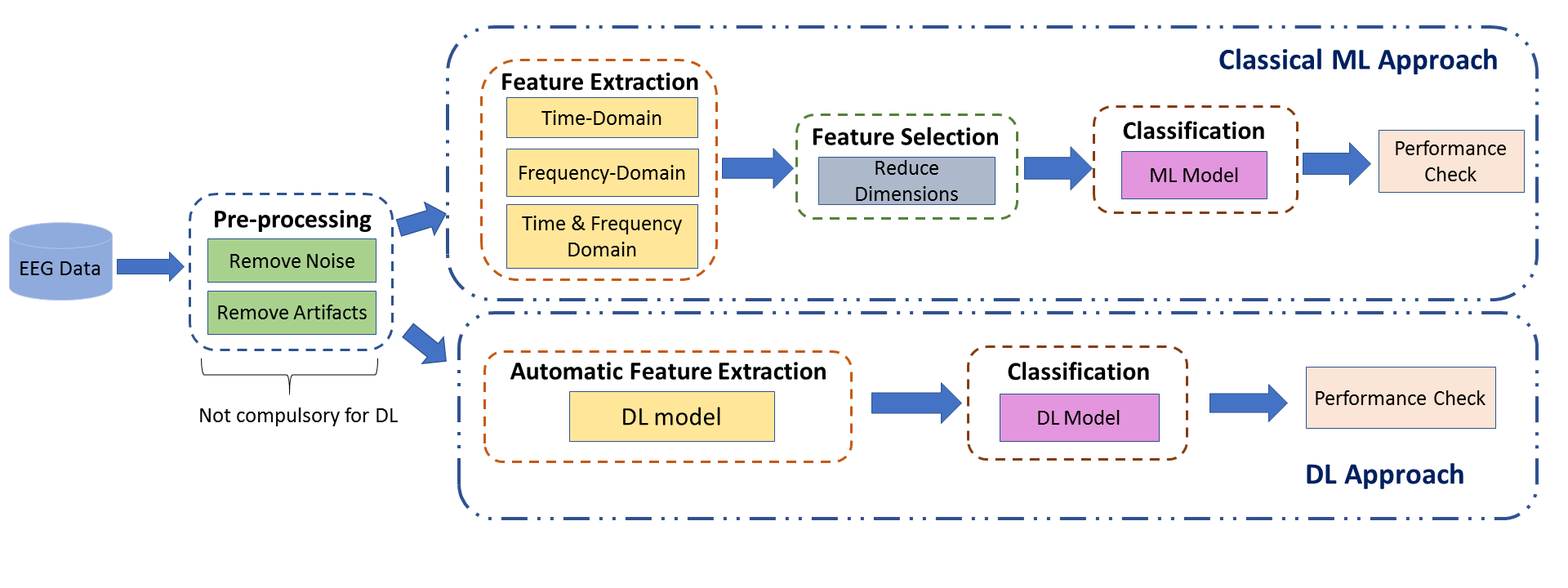}
\caption{Process of epilepsy prediction using EEG data and classification algorithm.}
\label{fig:ML_dataflow}
\end{figure*}

\subsubsection{Reinforcement Learning}

Animal psychology, how animals communicate with each other and with the environment, helped to develop reinforcement learning (RL) \cite{sutton1998introduction}. RL is a significant illustration of the advancement of technology due to the collaboration of neuroscience and AI. Reinforcement Learning is the process of developing a policy to maximize the rewards of interaction between an agent and its environment. Central factors of a reinforcement learning system are a policy, reward signal, value function, and model of the environment.

\subsection{ML for Epileptic Seizure Prediction}
Since the last century, researchers are working to overcome the hurdles related to the detection and prediction of epilepsy. As EEG signals are a key source for monitoring brain activity before, during, and after ES so the first focus of ES prediction research was on the analysis of EEG recordings. EEG signals are vitiated by eye-movements, blinks, cardiac signals, and muscle noise. Several filtering and noise reduction methods are used to decrease the effect of these various sources of noise and artifacts \cite{mannan2018identification}. After the removal of artifacts, significant features are needed for building ML models for the identification and classification of pre-ictal and interictal stages. Figure \ref{fig:ML_dataflow} shows the classical ML methodology for the epilepsy prediction and also highlight the major difference between the use of ML and DL technique. Basically one can give the raw data or minimally processed data (i.e., without extraction of features from the raw data) to a  DL model for pattern learning.

\subsubsection{Signal processing} 
Noise and artifact identification is a crucial procedure in raw biomedical signals. To reduce the influence of these artifacts in feature extraction, filtering of these artifacts is needed. Multiple techniques have been employed for filtering e.g. band-pass filter, wavelet filter, finite impulse response filter, and adaptive filter. This processing is also performed to normalize the data to make it comparable with the recording of other patients. There are also many data dropouts or corrupted data in the EEG recording due to limitations of implanted electrodes which lead to the insignificant performance of algorithms. Due to muscle artifacts and environmental noise, there also exist some outliers in data. The presence of these outliers badly influences the extracted features.

\subsubsection{Feature Extraction and Selection}
\begin{figure}[!htb]
\centering
\includegraphics[width=0.45\textwidth]{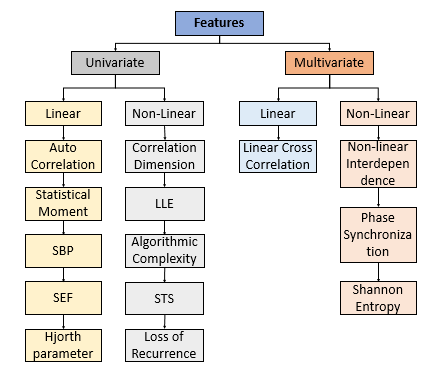}
\caption{Feature classification based on number of channels of EEG data.}
\label{fig:feats}
\end{figure}

All prediction models need reliable features, well correlated with pre-ictal and interictal stages. One can categorize these features based on the number of EEG channels as univariate (measures taken on each EEG channel separately) and multivariate (measures taken on two or more EEG channels) features. Further categorization of each of these is as linear or nonlinear features. Florian et al. compared the performance of univariate and bivariate measures containing both linear and non-linear strategies for ES prediction \cite{mormann2005predictability}. They noted that while using univariate measures, pre-ictal variations transpired 5-30 min before ES onset. While bivariate measures performed better by capturing pre-ictal changes at least 240 min before an ES onset. Fig \ref{fig:feats} shows some of the linear and nonlinear measures used in the literature for ES prediction. Linear measures performed better or some times similar to nonlinear measures.
 
\subsubsection{Classification} Identification of pre-ictal and interictal patterns from EEG data is carried out using ML algorithms, e.g, artificial neural network (ANN), k-means clustering, decision trees, SVM, and fuzzy logic. Mostly threshold-based on features values are utilized to make conclusions. However, ML-based studies broadly focused on the extraction of optimized features for prediction.

\begin{table*}[]
\centering
\caption{Summary of ML methods used for ES prediction. N/M indicates the not mentioned entries in the table.}
\label{tab:ML_sum}
\begin{tabular}{|l|l|l|l|l|l|l|l|l|l|}
\hline
\textbf{Year} & \textbf{Ref} & \textbf{Predictive characteristics} & \textbf{Model} & \textbf{\begin{tabular}[c]{@{}l@{}}EEG\\ Type\end{tabular}} & \textbf{\begin{tabular}[c]{@{}l@{}}No. of\\ Patients\end{tabular}} & \textbf{\begin{tabular}[c]{@{}l@{}}No. of Seizure\\ per Patient\end{tabular}} & \textbf{\begin{tabular}[c]{@{}l@{}}Prediction \\ Time\end{tabular}} & \textbf{Sensitivity} & \textbf{\begin{tabular}[c]{@{}l@{}}False\\ Positive per hr\end{tabular}} \\ \hline
\multicolumn{10}{|l|}{\textbf{MIT Database}} \\ \hline
2017 & \begin{tabular}[c]{@{}l@{}}Usman \\ et al. \cite{usman2017epileptic}\end{tabular} & \begin{tabular}[c]{@{}l@{}}Entropy, ApEn, HP, SpM,\\ SM\end{tabular} & SVM & Scalp & 24 & 3.5 & 23.48 min & 92.23\% & N/M \\ \hline
2018 & \begin{tabular}[c]{@{}l@{}}Usman \\ et al. \cite{usman2018efficient}\end{tabular} & \begin{tabular}[c]{@{}l@{}}Variance, Skewness, SD,\\ HP, Entropy, Kurtosis\end{tabular} & \begin{tabular}[c]{@{}l@{}}KNN\\ Naive Bayes\\ SVM\end{tabular} & Scalp & 24 & 3.5 & 34 min & \begin{tabular}[c]{@{}l@{}}97.44\%\\ 90.66\%\\ 97.07\%\end{tabular} & N/M \\ \hline
2018 & \begin{tabular}[c]{@{}l@{}}Kitano \\ et al. \cite{kitano2018epileptic}\end{tabular} & \begin{tabular}[c]{@{}l@{}}Zero-crossing of DWT\\ coefficients\end{tabular} & SOM & Scalp & 9 & \textgreater{}4 & N/M & 98\% & N/M \\ \hline
\multicolumn{10}{|l|}{\textbf{Freiburg Database}} \\ \hline
2017 & \begin{tabular}[c]{@{}l@{}}Sarif \\ et al. \cite{sharif2017prediction}\end{tabular} & \begin{tabular}[c]{@{}l@{}}Distribution of 6\\ fuzzy rules\end{tabular} & SVM & iEEG & 19 & 4.4 & 42 min & 96.6\% & 0.05-0.08 \\ \hline
2018 & \begin{tabular}[c]{@{}l@{}}Yang\\ et al. \cite{yang2018epileptic} \end{tabular} & Permutation Entropy & SVM & iEEG & 21 & \textgreater{}2 & 61.93 min & 94\% & 0.111 \\ \hline
\multicolumn{10}{|l|}{\textbf{EPILEPSIAE Database}} \\ \hline
2015 & \begin{tabular}[c]{@{}l@{}}Bandarabadi\\ et al. \cite{bandarabadi2015epileptic}\end{tabular} & \begin{tabular}[c]{@{}l@{}}Amplitude distribution\\ histogram \& Spectral power\end{tabular} & N/M & \begin{tabular}[c]{@{}l@{}}iEEG/E\\ EG\end{tabular} & 24 & 3.6 & 8 sec & 73.98\% & 0.06 \\ \hline
2017 & \begin{tabular}[c]{@{}l@{}}Direito\\ et al. \cite{direito2017realistic}\end{tabular} & 22 univariate features & SVM & \begin{tabular}[c]{@{}l@{}}iEEG/E\\ EG\end{tabular} & 216 & 5.6 & N/M & 38.5 & 0.2 \\ \hline
\multicolumn{10}{|l|}{\textbf{IEEG.org Database}} \\ \hline
2018 & \begin{tabular}[c]{@{}l@{}}Assi \\ et al. \cite{assi2018bispectrum}\end{tabular} & \begin{tabular}[c]{@{}l@{}}Bi-spectral Entropy\\ Bi-spectral Squared Entropy\\ Mean magnitude of\\ bispectrum\end{tabular} & MLP & iEEG & 3 Dogs & N/M & N/M & N/M & N/M \\ \hline
\end{tabular}
\end{table*}

\paragraph{Use of Bispectral Features to Predict Seizure }
Higher-order spectrum (HOS) features of iEEG recordings used to detect the seizure in earlier studies \cite{chua2009analysis}. However, Assi et al. \cite{assi2018bispectrum} used the HOS features to present that the bispectrum analysis of EEG provides significant phase information. They showed that the normalized bispectral entropy and the normalized squared bispectral entropy decreased during the pre-ictal state of seizure. They extracted these features from the 30 sec non-overlapping windows of iEEG recordings of epileptic dogs. They trained a 5-layer multilayer perceptron (MLP) for the classification of pre-ictal and interictal classes. The input layer of MLP consisted of 16 nodes as there are 16 channels of iEEG signals. They added 3 hidden layers of 30, 60 and 30 nodes of ReLu activation function. They computed the F1 score and p-value corresponding to pre-ictal and interictal distribution using each feature. However, researchers prefer to analyse the performance of the algorithm in terms of sensitivity and specificity for defined seizure prediction horizon and seizure occurrence period. This aspect is missing in this study.

Permutation entropy (PE) has been used in various early studies to characterize the EEG states of epilepsy \cite{mormann2003automated}, \cite{bruzzo2008permutation}. In 2007, Li et al. \cite{li2007predictability} used PE to distinguish pre-ictal states in rats. Recently, Yang et al. \cite{yang2018epileptic} used PE as a feature extracted from the iEEG data of Freiburg hospital data. They analyzed 83 seizures from 19 patients. They trained an SVM classifier with RBF kernel using 5 sec segments of features as input. Sensitivity and false prediction rate (FPR) used as performance analysis measures. They achieved 94\% sensitivity and 0.11 FPR on average with a mean SPH of 61 min.

\paragraph{Use of Selected Amount of Data to Predict Seizure}
To reduce the dimensions of EEG is one of the foremost concerns of researchers for the processing of data and predicting seizure using this data. Various dimension reduction techniques have been proposed with some pros and cons. In the resent work, Kitano et al. \cite{kitano2018epileptic} proposed the use of a small amount of data to predict seizure. They used only 20 min data of 9 patients out of the hours-long recording of 24 patients of CHBMIT database. 20 min data consisted of the 10 min pre-ictal and 10 min interictal data. They applied DWT on 4sec non-overlapping windows of this 20 min data and extracted zero-crossings of level 1 detailed coefficients of DWT. They used a self-organizing map (SOM), formerly introduced by Teuvo Kohonen in 1982 \cite{heskes1999energy}, for mapping the input data in clusters of pre-ictal and interictal states. They achieved 98\% sensitivity using the selected amount of data.

Although Kitano et al. achieved highly significant results on the selected amount of data, there are various flaws with this selection. Pre-ictal and interictal patterns have temporal variations across patients and with inpatient. To randomly select 10-min pre-ictal data is not a significant approach. Training of models on a small amount of data leads to the over-fitting of results. With the training on the randomly selected small amount of data, the model might not be able to show significant performance in real-time scenarios. Summary of recent work on ES prediction using ML techniques is in Table \ref{tab:ML_sum}.

\begin{table*}[]
\centering
\caption{Summary of DL methods used for ES prediction}
\label{tab:DL_sum}
\scalebox{0.92}{
\begin{tabular}{|l|l|l|l|l|l|l|l|l|l|}
\hline
\textbf{Year} & \textbf{Ref} & \textbf{\begin{tabular}[c]{@{}l@{}}Predictive \\ characteristics\end{tabular}} & \textbf{Database} & \textbf{\begin{tabular}[c]{@{}l@{}}EEG\\ Type\end{tabular}} & \textbf{\begin{tabular}[c]{@{}l@{}}No. of\\ Patients\end{tabular}} & \textbf{\begin{tabular}[c]{@{}l@{}}No. of Seizure\\ per Patient\end{tabular}} & \textbf{\begin{tabular}[c]{@{}l@{}}Prediction \\ Time\end{tabular}} & \textbf{Sensitivity} & \textbf{\begin{tabular}[c]{@{}l@{}}False\\ Positive/hr\end{tabular}} \\ \hline
\multicolumn{10}{|l|}{\textbf{CNN}} \\ \hline
2017 & \begin{tabular}[c]{@{}l@{}}Haider\\ et al. \cite{khan2017focal} \end{tabular} & Wavelet Transform & \begin{tabular}[c]{@{}l@{}}MSSM\\ CHB-MIT\end{tabular} & Scalp & 47 & 2.78 & \begin{tabular}[c]{@{}l@{}}8 min\\ 6 min\end{tabular} & 87.8\% & 0.142 \\ \hline
2018 & \begin{tabular}[c]{@{}l@{}}Truong\\ et al. \cite{truong2017generalised} \end{tabular} & STFT & \begin{tabular}[c]{@{}l@{}}Freiburg\\ CHB-MIT\\ American Epilepsy Society\end{tabular} & \begin{tabular}[c]{@{}l@{}}iEEG/\\ Scalp\end{tabular} & \begin{tabular}[c]{@{}l@{}}28 humans\\ 5 canines\end{tabular} & N/M & 5 min & \begin{tabular}[c]{@{}l@{}}81.4\%\\ 81.2\%\\ 82\%\end{tabular} & \begin{tabular}[c]{@{}l@{}}0.06\\ 0.16\\ 0.22\end{tabular} \\ \hline
2019 & \begin{tabular}[c]{@{}l@{}}Ramy Hussain\\ et al. \cite{hussein2019human} \end{tabular} & STFT & \begin{tabular}[c]{@{}l@{}}Melbourne seizure\\ prediction competition\\ dataset\end{tabular} & iEEG & 3 & 380 & 5 min & 87.8\% & N/M \\ \hline
\multicolumn{10}{|l|}{\textbf{LSTM}} \\ \hline
2018 & \begin{tabular}[c]{@{}l@{}}Tsiouris\\ et al. \cite{tsiouris2018long} \end{tabular} & \begin{tabular}[c]{@{}l@{}}Various time and\\ frequency features\end{tabular} & CHB-MIT & Scalp & 24 & 7.7 & 15-120 min & 99.28\% & 0.11-0.02 \\ \hline
\multicolumn{10}{|l|}{\textbf{GAN}} \\ \hline
2019 & \begin{tabular}[c]{@{}l@{}}Troung \\ et al. \cite{truong2019epileptic} \end{tabular} & STFT & \begin{tabular}[c]{@{}l@{}}Freiburg\\ CHB-MIT\\ EPILEPSIAE\end{tabular} & \begin{tabular}[c]{@{}l@{}}iEEG/\\ Scalp\end{tabular} & 56 & 6.8 & 5 min & N/M & N/M \\ \hline
\multicolumn{10}{|l|}{\textbf{DCAE + Bi-LSTM}} \\ \hline
2019 & \begin{tabular}[c]{@{}l@{}}Hisham\\ et al. \cite{daoud2019efficient}\end{tabular} & Raw data & CHB-MIT & Scalp & 8 & 5.37 & 1 hr & 99.72\% & 0.004 \\ \hline
\end{tabular}}
\end{table*}

\subsection{Introduction to DL For Healthcare}
DL models are the result of advancements in ML research that provide an ability to process raw data. DL models comprise of multiple layers of computational (non-linear) modules that work mutually to process data and produce an ultimate result. These multiple layers help in extraction of appropriate features and their examination or analysis for the output result. For example, in the classification task, higher layers of representation amplify features of the input that are significant for discrimination and subdue unnecessary variations. The core of DL models is that they contain modular layers that are designed to learned data using general-purpose algorithms \cite{lecun2015deep}. These layers are building blocks of deep neural networks (DNN). Commonly used neural networks are \textit{convolutional neural network} (CNN) and \textit{recurrent neural network} (RNN) \cite{lecun2015deep}. The structure of CNN is similar to that of the connectivity pattern of neurons in the brain. Convolution operation of a CNN is just like a filter with weights for extracting the features from multi-dimensional input data. While the RNNs are used to find logical sequences in input data. The output of each hidden layer passed to the next layer and also fed back to itself. Simply, the current output is a combined experience of the present moment and history. The key difference between CNN and RNN architecture is that CNNs only consider the current input while RNN considers current input and as well as the previous input, i.e., it contains memory logic. RNN performs significantly better on time series data while CNN is good for tasks like image classification. 


DL architectures have been used in many medical domains, e.g., in clinical imaging \cite{lee2017deep}, genomics, and proteomics \cite{asgari2015continuous}, computational biology \cite{angermueller2016deep}, and disease prediction \cite{khan2017focal}. DL algorithms are turned out to be adequate in detecting intricate patterns in high-dimensional data for classification, especially in EEG data. CNN is a widely used neural network for the training using EEG data because it can be very effective to reduce noise \cite{article}.

\subsection{DL for Neuroscience}
DL is solving problems in many fields, however, a potent relation exists between DNN and the study of the nervous system. ANNs were considered as a model for brain activity computations \cite{hassabis2017neuroscience}, while CNNs are models of visual information processing and the activations of hidden layers of CNN are considered as the activity of neurons in connected brain regions associated with the processing of visual sensory motors. Deep networks are a valuable mean of computation in neuroscience as these are statistical time-series models of neural activity in the brain, e.g, the CNN can act as an encoding model of computational neuroscience. In connectomics, to understand the mapping of the connectivity of neural networks in the brain, deep networks are used to understand the connectivity of neural units from 3D electron microscopic images \cite{jain2010machines}. The existing era of advancement is accelerating the research of neuroscience-inspired ML tools \cite{hassabis2017neuroscience}.

\subsection{DL for ES Prediction}
ML classification algorithms use feature vectors, derived from traditional signal processing methods for training and provide good accuracy but a generalized model can not be anticipated from these techniques. For seizure prediction through an ML approach, script writing requires feature extraction stage that takes a lot of time. The presence of noise and artifacts in data makes feature extraction very complex to handle. Hence it is a challenging problem to produce a generalized automatic system with loyal performance especially even when limited training samples are available. On the other hand, DL algorithms automatically learn features and give encouraging outcomes in ES prediction. Features learned through DL models are more distinguishing and robust than hand-crafted features \cite{antoniades2016deep}. 

\begin{table*}[]
\centering
\caption{Overview of EEG databases}
\label{tab:database}
\scalebox{0.96}{
\begin{tabular}{|l|l|l|l|l|l|l|}
\hline
\textbf{Database} & \textbf{No. of Subjects} & \textbf{No. of Channels} & \textbf{Recording Type} & \textbf{No. of ES} & \textbf{\begin{tabular}[c]{@{}l@{}}Duration of Each \\ Recording (Hour)\end{tabular}} & \textbf{\begin{tabular}[c]{@{}l@{}}Sampling Frequency\\    (Hz)\end{tabular}} \\ \hline
CHB-MIT & 24 & 23 & Scalp EEG & 198 & \begin{tabular}[c]{@{}l@{}}1 (some cases have \\ 2-4 hours of recording)\end{tabular} & 256 \\ \hline
MSSM & 28 & 22 & Scalp & 61 & 48-192 & 256 \\ \hline
Freiburg & 21 & 128 & iEEG & 88 & At least 24 & 256 \\ \hline
Bonn & \begin{tabular}[c]{@{}l@{}}25 (5 sets each\\ consists of recording\\ of 5 subjects)\end{tabular} & \begin{tabular}[c]{@{}l@{}}1 (100 flies of single\\ channel data in each\\ set)\end{tabular} & Scalp/iEEG & \begin{tabular}[c]{@{}l@{}}Dataset E is the\\ recording\\ of ictal stage\end{tabular} & 23.6 & 173 \\ \hline
EPILEPSIAE & 30 & 122 & Scalp/iEEG & 1800+ & 96 & 250-2500 \\ \hline
TUH & 10874 & 24-36 & iEEG & $\approx$ 14777 &-& 250\\ \hline
\end{tabular}
\label{tab:datasets}}
\end{table*}



\subsubsection{Use of CNN for ES Prediction}
To introduce a method that can be applied for all patients with minimum pre-processing of EEG data Troung et al. \cite{truong2017generalised} proposed a CNN based prediction method. They used the Freiburg hospital iEEG database and CHBMIT scalp EEG database for training and testing of the CNN model. Short-term Fourier transform (STFT) used to transform the raw EEG data into a two-dimensional matrix. This image is then fed to the CNN for feature learning and classification of pre-ictal and interictal states. For evaluation of the performance of the algorithm, they set the seizure prediction horizon (SPH) to 5 min and seizure occurrence period (SOP) to 30 min and used sensitivity and false prediction rate as evaluation metrics. While following a leave-one-out cross-validation, they reached 79.7\% sensitivity with 0.24 FPR on raw EEG and 89.8\% sensitivity with 0.17 FPR on standardized data.

For real-time clinical use of ES predictor, SPH must be long enough to allow the patient to come out of a dangerous situation and take precautionary measures and SOP should not be too long. The work of Haider et el. \cite{khan2017focal} performed better than previous work by giving 87.8\% sensitivity and 0.142 FPR with 10 min SPH. They used CHBMIT and MSSM databases for the training and the testing of the model. Raw EEG converted into wavelet tensors and CNN used to extract features from transformed data for classification of pre-ictal and interictal data.

After the establishment of the feasibility of ES prediction in a clinical setting by demonstrating the success of implantable recording system by Cook et al. \cite{cook2013prediction}, new avenues of further research have been opened. To take the work of Cook et al. forward, Isabell et al. \cite{kiral2018epileptic} presented a portable seizure prediction system with tunable parameters according to the patient's need. They transformed iEEG data into spectrograms and used frequency transformed data as an input to the deep learning model for automatic pre-ictal feature learning. These tunable parameters are the sensitivity of the system, duration, and the number of alarms. For the tuning of these parameters, the authors added a processing layer in the model. They deployed their prediction algorithm on a low-power TrueNorth chip to introduce a wearable device. Their prediction system performed an average sensitivity of 69\% and average time in warning of 27\%, significantly exceeding a comparable random predictor for all patients by 42\%.

Motivated by the work of Cook et al. \cite{cook2013prediction} and Karoly et al. \cite{karoly2016interictal}, demonstrating that the seizure prediction algorithm could not produce satisfactory prediction sensitivity for some patients, Ramy Hussain et al. \cite{hussein2019human} worked on some part of the data of these patients. They applied a technique of downsampling to reduce the dimension of data by a factor of 4. They explained that handcrafted features are not suitable for authentic ES prediction because the EEG data not only varies between patients but also varies for the same patient over time. They transformed the EEG data by applying STFT and fed this transformed data to CNN. To learn local features they used 1x1 convolutional layers and for abstract feature learning, they used larger convolutions. They obtained 87.85\% sensitivity and 0.84 AUC on average as a performance measure of the prediction algorithm. They also explained that reasons for the limited performance of ES prediction algorithms are data drop-outs, data mismatch, imbalance distribution and outliers in data.

\subsubsection{Unsupervised DL Method for ES Prediction}
One problem in the ES prediction is the availability of labeled data. To overcome this problem the first step taken by Troung et al. \cite{truong2019epileptic} is the use of a generative adversarial network (GAN) to do unsupervised training. They fed the spectrogram of STFT of EEG to GAN and used trained discriminator as a feature to predict seizures. This unsupervised training is significant because it not only provides real-time prediction using EEG recording also does not require manual effort for feature extraction. They used AUC as a performance measure with 5 min SPH and SOP of 30 min. They compared their results with supervised methods of model training, and their approach performed well with 77.68\% AUC for CHBMIT scalp EEG data, 75.47\% AUC on Freiburg hospital data and 65.05\% AUC with EPILEPSIAE database. A summary of these works is presented in Table \ref{tab:DL_sum}.
 
 \subsubsection{Use of RNN for ES Prediction}
 
Tsiouris et al. \cite{tsiouris2018long} used long short-term memory (LSTM) for the first time for the prediction of an epileptic seizure. They compared the performance of different architectures of LSTM for randomly selected input segment size of 5-50 sec. They compared the performance of three architectures of LSTM using feature vectors of EEG segments as input to LSTM, where the feature vector consists of various features from the time domain, frequency domain, and local and global measures from graph theory. LSTM-1 architecture consisted of a single layer with 32 memory units, while the number of memory units increased to 128 in LSTM-2 architecture. The number of memory units preserved at 128 but an extra layer of equal dimension added to LSTM-3. The performance of LSTM-3 was the best among the three considered architectures. Using the pre-ictal window of 15 min, they also evaluated the performance of LSTM-3 for raw EEG as input as compared to the performance of the feature vector. They showed that deep architecture with raw EEG input and satisfactory performance is still an open issue in the said field. On average, LSTM-3 performed better with 99.28\% sensitivity for 15 min pre-ictal period, 99.35\% sensitivity for 30 min pre-ictal period, 99.63\% sensitivity with 60 min pre-ictal period, and 99.84\% sensitivity with 120 min pre-ictal period. However, too much feature engineering needed for these results.

\subsection{Datasets used for ES Prediction and Analysis} 
EEG is becoming a prevailing mean of acquiring brain signals to detect and predict ES. To this end, various open-access databases have been published by various hospitals and research centers. For instance, the Center of Epilepsy at Children's Hospital, Boston and Temple University Hospital have made their EEG databases publicly available to the researchers who aim to develop ML/DL models and other statistical analysis based methods \url{physionet.org}. In Table \ref{tab:datasets}, we provide a summary of such widely used and publicly available databases. Although, the database of Bonn University is not large enough but is extensively used for the detection of ES in the literature. It consists of 5 datasets A, B, C, D and E. CHB-MIT database has data of 22 patients with 9-24 recordings of each patient and every recording is 1 hour long with some discontinuities due to hardware limitation (some cases have 2-4 hours long recordings). Freiburg Hospital's database was one of the considerable databases which contained iEEG data of 21 subjects with around 88 seizures but recently it has been merged into EPILEPSIAE database to provide more larger datasets due to which this database is not open-source now.


\subsection{Evaluation Metrics for ES Prediction} 

The clinical employment of ES prediction methods requires a sufficient performance and quality check and different evaluation metrics have been proposed in the ES prediction literature. For instance, Osorio et al. proposed sensitivity and false prediction rate as performance parameters of ES predictors \cite{osorio1998real}. Sensitivity is measured as the ratio of correctly predicted seizures to all seizures. Moreover, contrary to the ideal situation, one can not prevent false prediction and with the increase in sensitivity, the false prediction rate also increases. The widely used evaluation metrics are described below. 

\begin{equation*}
  Accuracy=\dfrac{TP+TN}{TP+TN+FP+FN}
\end{equation*}

\begin{equation*}
 Sensitivity=\dfrac{TP}{TP+FN}
\end{equation*}

\begin{equation*}
 Specificity=\dfrac{TN}{TN+FP}
 \end{equation*}
Where 
\begin{itemize}
\item True positive (TP) is the number of correctly predicted ES.
\item False Negative (FN) is the number of ES that are incorrectly predicted as not seizures. 
\item True negative (TN) is the number of correctly predicted no-seizure. 
\item False Negative (FP) is the number of non-ES that are incorrectly predicted as seizures. 
\end{itemize}

\begin{figure}[!htb]
\centering
\includegraphics[width=0.45\textwidth]{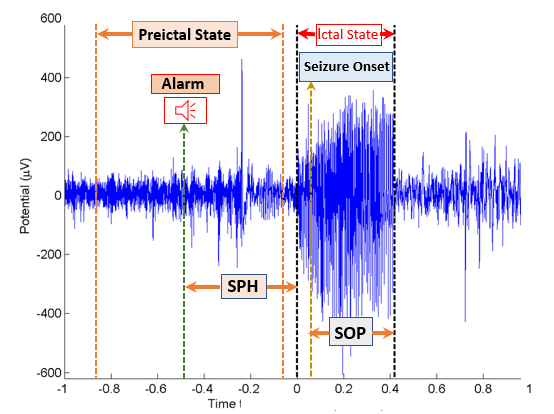}
\caption{Concept of seizure occurrence period (SOP) and Seizure prediction horizon (SPH). With a precise prediction, a seizure must occur after SPH and within the SOP.}
\label{fig:seizure period}
\end{figure}

An ES predictor generates an alarm before seizure onset and according to ideal situation the predictors must anticipate the exact time of onset. In practical applications, a predictor anticipates a duration of the high probability of occurrence of seizure. So, another performance check metric is a \textit{seizure occurrence period} (SOP), the time duration in which there is a possibility of seizure. Another metric is the \textit{seizure prediction horizon} (SPH), the duration of time between the alarm and the inception of SOP \cite{maiwald2004comparison}, \cite{snyder2008statistics}. In Fig \ref{fig:seizure period} the concept of SPH and SOP is illustrated.

\section{Pitfalls of Epileptic Seizure Prediction Methods}
\label{sec:pitfalls}

ML has solved several challenges for ES prediction that include manual, tedious, and time-consuming analysis methods. Model interpretation is crucial and pattern identification in data is as significant as data fitting. A fundamental difficulty in bio-medicine is the correct classification of ailment and its sub-types. Enormous available biomedical data can lead to the identification of more comprehensive sub-types. One can easily find various ways in which ML, specifically DL has improved the EEG analysis. The hierarchical nature of neural networks has significantly developed the potential of learning features from raw data or  minimally processed data. Automatically learned features through DL models are more powerful and effective than those extracted by analytical tools. This shows that DL has the potential to give high performance on analysis tasks. Research on epilepsy prediction has been going for many years and much progress has been made using different approaches, but there also exist many problems. Some potential pitfalls related to ES prediction using ML methods are discussed next.

\subsection{Unavailability of Open Access EEG Data}
A core problem in the ES prediction and analysis research is the unavailability of long-term EEG data. In 2005 Iasemidis et al. performed the prediction alarm almost 91 min before the ES onset on private EEG data \cite{iasemidis2005long}. However, no one has been able to reproduce these results since then on any publicly open EEG data. so, there is an urgent need for open access sharing of EEG databases with long-term recordings and also code sharing (using Github or similar repositories) for reproducibility of findings.

\subsection{Data Dropouts}
One of the main reasons for the low performance of prediction algorithms is the missing observations. There are many zero or nearly equal to zero values in the observed data because of the failure of communication between the wearable devices or implanted devices with limited storage capacity and storage device for several possible reasons. Learning from corrupt, or missing, data has not attained much attention to the machine learning community. However, there is a need to add missingness indicators in models that can provide significant pieces of information for making predictions.


\subsection{Incapability to Predict ES Using Raw Signals}

Time consumption and computational cost will increase due to excessive feature extraction. We require a quick prediction with comparatively low-power hardware and cheaper computational cost so that the real-time system for ES prediction becomes feasible. Unfortunately, researchers are not able to build a model for learning feasible features from raw signals yet. Although DL has greatly rectified the problem of feature extraction by automatically extracting features from pre-processed data, the limitation is that these methods require an abundant amount of data for effective prediction \cite{ghassemi2018opportunities}.

\subsection{Lack of Efficient Hardware Applications}
The main aim of ES prediction research is to improve the patients' quality of life. In this review, we have provided a comprehensive assessment of the state-of-the-art in ML as applied to ES prediction however the cost-effective and efficient hardware implementation remains the bottleneck. Although some initial work on hardware implementation has been performed \cite{dilorenzo2019neural} but much work is required as to how cost-effective devices can be manufactured which produce optimized results and how can their use be made common? And it is only then can the patients benefit from this research.

\section{Future Directions and Open Research Issues}
\label{sec:insights}

In this section, we present directions for future work in ES prediction and various open research issues that require further investigation. 



\subsection{Curse of Data Dimensionality}
EEG signals are recorded using multiple electrodes due to which the dimensions of the recorded signals increases and analyzing multi-channel EEG signal become difficult. An ideal approach is to convert multi-channel EEG data into a single channel by applying appropriate signal analysis techniques (e.g., converting them to spectrograms) or to make use of the single-channel EEG signals from the collection of brain signals. It has been observed that interesting seizure-related brain activities were weak in a few signals of multi-channel EEG \cite{wang2018adaptive}. Therefore, the selection of a good quality signal for effective prediction of seizure is crucial. In a recent study, signal quality index (SQI) based adaptive algorithm is presented for best channel selection in multi-channel EEG of nonconvulsive seizure patient \cite{wang2018adaptive} and substantial efforts have been made on developing adaptive algorithms for EEG channel selection using different dimensionality reduction techniques such as principal component analysis (PCA) \cite{birjandtalab2017automated}, independent component analysis (ICA) \cite{viola2010using}, and discernibility matrix-based dimensionality reduction \cite{chatterjee2016discernibility}, etc. However, the development of an optimal strategy is still an open research problem. 


\subsection{Data Annotation} 
In the literature, the problem of ES prediction and detection is mostly formulated as a supervised learning task that requires labeled data. The EEG recordings are manually annotated by expert neurologists and physicians which is a costly, time-consuming, and tidy task. The performance of the ML techniques significantly depends upon the quality of annotations and to increase the efficacy of ML techniques, and in particular DL the natural approach is to use more training data. The development of a true validation set for assessing the performance of the trained model is also important. However, the annotation of large-scale collections of EEG recordings into respective categories is practically not feasible, it hinders the applicability of ML/DL techniques. This necessitates the development of automated ways for data labeling such as active learning, data labeling using generative models (for instance, ES prediction using GAN is presented in \cite{truong2019epileptic}), and unsupervised clustering-based classification, etc. 



\subsection{Distribution Data Sharing and Management}
In clinical settings, patients' data is produced across different facilities and to develop efficient ML/DL techniques, sharing of distributed data across different departments and as well across different hospitals is required. Moreover, data from different domains can be integrated to extract knowledge required for different tasks, e.g., the annotation. Recurrent models and different natural language processing (NLP) techniques can be used to extract rich knowledge from raw clinical notes and electronic health records (EHR) that can enhance the capability of data annotators. In addition, ML/DL models can be developed that are capable of learning from heterogeneous sources and distributed data. However, the cost of data sharing and management can be huge and also this will lead to new challenges of data integrity, availability, and privacy. This direction of work requires innovative ways to encourage data pooling and sharing. 



\subsection{Interpretable ML}
Despite the state of the art performance of DL techniques, these methods are black-box models and lack underlying theory about their learning behavior and thought process. Therefore, their decisions are not interpretable due to which uncertainty quantification of predictions becomes extremely difficult. In addition, the life-critical nature of healthcare applications demands that DL models' decision should be explainable and interpretable at the same time. It has been argued that interpretable models enable the extraction of most relevant and important features for the specific tasks for which they are developed \cite{murdoch2019interpretable}. In a recent study, a visualization framework named Deep-Tune is presented that enables neuroscientists to identify patterns that activate a certain neuron in a CNN model that was trained for the task of neural spike rate prediction \cite{abbasi2018deeptune}. The work on developing interpretable ML models is catching up and is still an open research problem.

\section{Conclusions}
\label{sec:con}
In this paper, we comprehensively reviewed the available literature and highlighted why early prediction of ES is required, how ML and DL techniques are used for ES prediction. In the context of EEG analysis, feature selection, ES detection, and prediction, and the evaluation of prediction or detection algorithms, ES prediction is a capacious topic. Contrary to the findings of this paper, most of the previous survey papers focused only on EEG analysis and a few of them covered the developments of prediction techniques; while we tried to provide insights by considering the aspects of the feature selection, prediction techniques, and evaluation methodologies, etc. In addition, we have also highlighted future work directions and open research problems that require further investigation.


\bibliographystyle{unsrt}

\end{document}